\documentclass[runningheads,a4paper]{llncs}

\usepackage{amssymb}
\usepackage{amsmath}
\usepackage{bm}
\usepackage{color}
\definecolor{cobalt}{rgb}{0,0.28,0.28}

\pdfoutput=1
\usepackage{hyperref}
\hypersetup{
    colorlinks=true,
    linkcolor=cobalt,
    citecolor=cobalt,
    filecolor=cobalt,
    urlcolor=cobalt,
}

\usepackage{graphicx}
\usepackage{float}
\usepackage{pdfcomment}
\usepackage[capitalise]{cleveref}

\usepackage{url}
\urldef{\mailsa}\path|chloe-agathe.azencott@mines-paristech.fr|
\newcommand{\keywords}[1]{\par\addvspace\baselineskip
\noindent\keywordname\enspace\ignorespaces#1}

\newcommand{\eset}{\mathcal{E}}
\newcommand{\sset}{\mathcal{S}}
\newcommand{\tset}{\mathcal{T}}
\newcommand{\vset}{\mathcal{V}}

\newcommand{\vvec}{{\bm v}}
\newcommand{\yvec}{{\bm y}}
\newcommand{\betavec}{{\bm \beta}}

\DeclareMathOperator*{\argmax}{arg\,max}
\DeclareMathOperator*{\argmin}{arg\,min}

\newcommand{\lonenorm}[1]{\left|\left|#1\right|\right|_1}
\newcommand{\ltwonorm}[1]{\left|\left|#1\right|\right|_2}

\begin{document}

\mainmatter  

\title{Network-Guided Biomarker Discovery}

\titlerunning{Network-Guided Biomarker Discovery}

%
%
\author{Chlo\'e-Agathe Azencott\inst{1}}
\authorrunning{Azencott}

\institute{MINES ParisTech, PSL-Research University, CBIO-Centre for Computational Biology, 35 rue St Honor\'e 77300 Fontainebleau, France\\
Institut Curie, 75248 Paris Cedex 05, France\\
INSERM, U900, 75248 Paris Cedex 05, France
\mailsa}

\toctitle{Network-Guided Biomarker Discovery}
\tocauthor{Azencott}
\maketitle

\begin{abstract}
Identifying measurable genetic indicators (or biomarkers) of a specific condition of a biological system is a key element of precision medicine. 
Indeed it allows to tailor diagnostic, prognostic and treatment choice to individual characteristics of a patient. 
In machine learning terms, biomarker discovery can be framed as a feature selection problem on whole-genome data sets.
However, classical feature selection methods are usually underpowered to process these data sets, which contain orders of magnitude more features than samples.
This can be addressed by making the assumption that genetic features that are linked on a biological network are more likely to work jointly towards explaining the phenotype of interest.
We review here three families of methods for feature selection that integrate prior knowledge in the form of networks.

\end{abstract}

\section{Introduction and Motivation}
\label{introduction}

Therapeutic development today is largely based on large-scale clinical trials and the average responses of thousands of people.
However, a large number of medical conditions have no satisfactory treatment, and when treatment is available, many patients either do not respond or experience unacceptable side effects~\cite{spear2001}.
This is explained both by variations in environment and life styles between individuals, and by their genetic differences. 
As a consequence, precision medicine, which aims at tailoring preventive and curative treatments to patients based on their individual characteristics, is gaining considerable momentum. 
At its core, it relies on identifying features, genetic or otherwise, that correlate with risk, prognosis or response to treatment. 
Here we are interested in the identification, from large whole-genome dataset, of {\em genetic} features associated with a trait of interest.
Such features, which can be used to aid diagnostic, prognostic or treatment choice, are often refered to as {\em biomarkers}.
\par Biomarker discovery, which can be framed as a {\em feature selection} problem, depends on collecting considerable amounts of molecular data for large numbers of individuals.
This is being enabled by thriving developments in genome sequencing and other high-throughput experimental technologies, thanks to which it is now possible to accumulate tens of millions of genomic descriptors (such as single-nucleotide polymorphisms or copy number variations of the DNA, gene expression levels, protein activities, or methylation status) for thousands of individuals~\cite{reuter2015}.
However, these technological advances have not yet been accompanied by  similarly powerful improvements in the methods used to analyze the resulting data~\cite{vanallen2013}. 
\par One of the major issues we are facing is that feature selection methods suffer from a small sample size problem: they are statistically underpowered when the dimensionality of the data (the number of biomarkers to investigate) is orders of magnitude larger than the number of samples available.
This is one of the reasons behind the relative failure of genome-wide association studies to explain most of the genetic heredity of many complex traits~\cite{manolio2009}. 
\par This problem can be addressed by using {\it prior biological knowledge}, which reduces the space of possible solutions and helps capturing relevant information in a statistically sound fashion. 
When a human expert is available, this is a typical application case for interactive machine learning~\cite{holzinger2016}, where a domain expert drives a heuristic procedure to reduce the complexity of the search space.
This type of ``doctor-in-the-loop'' approach has recently been successfully applied in the clinic
\cite{hund2016}.
However, such analyses are currently restricted to relatively small numbers of a variables ($61$ in the example cited above) and it is not always possible to involve an expert directly. Hence, we will focus on using prior knowledge compiled in databases.
\par Because genes do not work in isolation, but rather cooperate through their interaction (physical, regulatory, or through co-expression) in cellular pathways and molecular networks, this prior biological knowledge is often available in a structured way, and in particular under the form of networks. Examples include the STRING database~\cite{szklarczyk2015}, which contains physical and functional interactions, both computationally predicted and experimentally confirmed, for over $2\,000$ organisms, or BioGRID~\cite{chatraryamontri2015}, which includes interactions, chemical associations, and post-translational modifications from the literature. In addition, systems biologists are building specialized networks, focused on the pathways involved in a particular disease. One example of such networks is ACSN~\cite{kuperstein2015}, a comprehensive map of molecular mechanisms implicated in cancer. These gene-gene interaction networks can be used to define networks between genomic descriptors, by mapping these descriptors to genes, using for instance in the case of SNPs a fixed-size window over the genetic sequence, and connecting together all descriptors mapped to the same gene, and all descriptors mapped to either of two interacting genes~\cite{azencott2013}.
We will here make the assumption that genetic features that are linked on such a network are more likely to work jointly towards explaining the phenotype of interest, and that such effects would otherwise be missed when considering them individually.
\par This chapter focuses on methods for feature selection that integrate prior knowledge as networks. Compared to pathway-based approaches, which assess whether predefined sets of genes are associated with a given trait, network-based approaches introduce flexibility in the definition of associated gene sets. We will review three families of approaches, namely post-hoc analyses, regularized regression and penalized relevance, before presenting their multi-task versions and discussing open problems and challenges in network-guided biomarker discovery.

\section{Glossary and Key Terms}
\label{glossary}


\textit{Feature selection:} In machine learning, feature selection~\cite{guyon2003} aims at identifying the most important features in a data set and discarding those that are irrelevant or redundant. This framework is clearly well-suited to the identification of biologically relevant features.\\[2ex]
\textit{Sparsity:} A model is said to be sparse when it only contains a small number of non-zero parameters, with respect to the number of features that can be measured on the objects this model represents~\cite{hastie2015}. This is closely related to feature selection: if these parameters are weights on the features of the model, then only the few features with non-zero weights actually enter the model, and can be considered selected. 
\\[2ex]
\textit{Genome-Wide Association Study (GWAS):} GWAS are one of the prevalent tools for detecting genetic variants associated with a phenotype.
They consist in collecting, for a large cohort of individuals, the alleles they exhibit across of the order of $250,000$ to several millions of Single Nucleotide Polymorphisms (SNPs), that is to say, individual locations across the genome where nucleotide variations can occur. 
The individuals are also phenotyped, meaning that a trait of interest (which can be binary, such as disease status, or continuous, such as age of onset) is recorded for each of them.
Statistical tests are then run to detect associations between the SNPs and the phenotype. 
A recent overview of the classical GWAS techniques can be found in~\cite{bush2012}. 
\\[2ex]
\textit{Graph / Network:} A graph (network) $(\vset, \eset)$ consists of a set of vertices (nodes) $\vset$ and a set of edges (links) $\eset$ made of pairs of vertices. If the pair is ordered, then the edge is directed; otherwise, it is undirected. A graph with no directed edge is called undirected; unless otherwise specified, this is the type of graph we consider here. We use the notation $i \sim j$ to denote that vertex $i$ and vertex $j$ form an edge in the graph considered. 
\\[2ex]
\textit{Adjacency matrix:} Given a graph $(\vset, \eset)$, its adjacency matrix is a square matrix $W \in \mathbb{R}^{d \times d}$, where $d = |\vset|$ is the number of vertices, and $W_{ij} \neq 0$ if and only if there is an edge between the $i$-th and the $j$-th elements of $\vset$. $W_{ij} \in \mathbb{R}$ represents the weight of edge $(i, j)$. If all non-zero entries of $W$ are equal to $1$, the graph is said to be unweighted.
\\[2ex]
\textit{Network module:} Given a graph $G = (\vset, \eset)$, a graph $G' = (\vset', \eset')$ is said to be a subgraph of $G$ if and only if $\vset'$ is a subset of $\vset$ and $\eset'$ is a subset of $\eset$. In systems biology,  the term ``network module'' refers to a subgraph of a biological network whose nodes work together to achieve a specific function. Examples of modules include transcriptional modules, which are sets of co-regulated genes that share a common function, or signaling pathways, that is to say chains of interacting proteins that propagate a signal through the cell. In the context of biomarker discovery, we are interested in finding modules of a given biological network that are associated with the phenotype under study. 
\\[2ex]
\textit{Graph Laplacian:} Given a graph $G$ of adjacency matrix $W \in \mathbb{R}^{d \times d}$, the Laplacian~\cite{merris1994} of $G$ is defined as $L:= D - W$, where $D$ is a $d \times d$ diagonal matrix with diagonal entries $D_{ii} = \sum_{j=1}^d W_{ij}$. 
The graph Laplacian is analog to the Laplacian operator in multivariable calculus, and similarly measures to what extent a graph differs at one vertex from its values at nearby vertices. Given a function $f: \vset \mapsto \mathbb{R}$, $f^\top L f$ quantifies how ``smoothly'' $f$ varies over the graph~\cite{smola2003}.
\\[2ex]
\textit{Submodularity:}
Given a set $\vset$, a function $\Phi: 2^\vset \rightarrow \mathbb{R}$ is said to be submodular if for any $\sset, \tset \subseteq \vset$, $\Phi(\sset) + \phi(\tset) \geq \Phi(\sset \cup \tset) + \Phi(\sset \cap \tset)$. This property is also referred to as that of diminishing returns. 
Given a graph $G$ and its adjacency matrix $W$, an example of submodular function is the function $\Phi: \sset \mapsto \sum_{p \in \sset} \sum_{q \notin \sset} W_{pq}$.
In the case of equality, i.e.  $\Phi(\sset) + \phi(\tset) = \Phi(\sset \cup \tset) + \Phi(\sset \cap \tset)$ for any $\sset, \tset \subseteq \vset$, $\Phi$ is said to be {\em modular}. 
In this case, the value of $\Phi$ over a set is equal to the sum of its values over items of that set.
The cardinality function $\Phi: \sset \mapsto |\sset|$ is a simple example of a modular function.
Submodular functions play an important role in optimization~\cite{fujishige2005} and machine learning~\cite{bach2013}.

\section{State of the Art}
\label{state-of-the-art}
\subsection{Network-based post-analysis of association studies}
\label{sec:post_hoc}
We start by describing methods that have been developed for the network-based analysis of GWAS outcomes; these methods can easily be extended to other type of biomarkers.
These approaches start from a classical, single-SNP GWAS, in which the association of each SNP with the phenotype is evaluated thanks to a statistical test. 
This makes it possible to leverage state-of-the-art statistical tests that, for example, account for sample relatedness~\cite{thornton2015}, address issues related to correlation between markers (linkage disequilibrium)~\cite{liu2013}, or are tailored to the discovery of rare variants~\cite{lee2014}. 
In addition, they can easily be applied without access to raw data, only on the basis of published summary statistics.
Their goal is to find modules of a given gene-gene network that concentrate more small $p$-values than would be expected by chance. 

The first step is to map all SNPs from the dataset to genes, and to summarize the $p$-values of all SNPs mapped to a given gene as a unique gene $p$-value. 
This summary can be based for example on the minimum, maximum, or average $p$-value. A popular alternative consists in using VEGAS, which accounts for linkage disequilibrium between markers~\cite{liu2010}. 

Several search methods have been proposed to find modules of significantly associated genes from such data. In dmGWAS~\cite{jia2012}, the authors use a dense module searching approach~\cite{chuang2007} to identify modules that locally maximize the proportion of low $p$-value genes. 
This search algorithm is greedy.
It considers each gene in the network as a starting seed, from which it grows modules by adding neighboring genes to the set as long as adding them increases the module's score by a given factor.\\
An alternative approach, first proposed in~\cite{baranzini2009} and refined in PINBPA~\cite{wang2015}, relies on a simulated annealing search called JActiveModule and first proposed for the discovery of regulatory pathways in protein-protein interaction networks~\cite{ideker2002}.\\
Finally, GrandPrixFixe~\cite{tasan2015} uses a genetic algorithm for its search strategy.

\paragraph{Limitations} 
Because exact searches are prohibitively expensive in terms of calculations, these approaches rely on heuristic searches that do not guarantee that the top-scoring module is found. Let us note however that any highly scoring module that is detected with such an approach is bound to be, if not biologically, at least statistically interesting. Methods exist to identify top-scoring sub-networks exactly, but they are too computationally intensive to have been applied to GWAS at this point~\cite{mitra2013}. An other way to mitigate this issue is to predefine potential modules of interest~\cite{akula2011}, but this strongly limits the flexibility offered by the use of networks rather than of predefined gene sets. Finally, these computational issues limit their application to networks defined over genes rather than directly over biomarkers.\\
More importantly, such methods rely on single-locus association studies, and are hence unsuited to detect interacting effects of joint loci. 
The failure to account for such intearcting effects is advanced as one of the main reasons why classical GWAS often does not explain much of the heritability of complex traits~\cite{marchini2005,manolio2009}.

\subsection{Regularized linear regression}
\label{sec:regularized_regression}
So-called embedded approaches for feature selection~\cite{guyon2003} offer a way to detect combinations of variants that are associated with a phenotype. 
Indeed, they learn which features (biomarkers here) contribute best to the accuracy of a machine learning model (a classifier in the case of case/control studies, or a regressor in the case of a quantitative phenotype) while it is being built.

\paragraph{Regularization}
Within this framework, the leading example is that of linear regression~\cite{tibshirani1994}.
Let us assume the available data is described as $(X, \yvec) \in \mathbb{R}^{n \times m} \times \mathbb{R}^{n}$, that is to say as $n$ samples over a $m$ biomarkers ($X$), together with their phenotypes ($\yvec$). 
A linear regression model assumes that the phenotype can be explained as a linear function of the biomarkers:
\begin{equation}
  \label{eqn:linreg}
  y_i = \sum_{p=1}^m X_{ip} \beta_p + \epsilon_i ,
\end{equation} 
where the regression weights $\beta_1, \dots, \beta_m$ are unknown parameters and $\epsilon_i$ is an error term. Note that we can equally assume that the mean of $y$ is $0$, or that the first of the $m$ biomarkers is a mock feature of all ones that will serve to estimate the bias of the model. The least-squares methods provides estimates of $\beta_1, \dots, \beta_m$ by minimizing the least-square objective function (or data-fitting term) given in matrix form by Eq.(\ref{eqn:lse}):
\begin{equation}
  \label{eqn:lse}
  \argmin_{\beta \in \mathbb{R}^m} \ltwonorm{X \betavec - \yvec}^2.
\end{equation} 

When $m \gg n$, as it is the case in most genome-wide biomarker discovery datasets, Eq.(\ref{eqn:lse}) has an infinite set of solutions. 
In order to {\em regularize} the estimation procedure, one can add to the least-square objective function a {\em penalty term}, or {\em regularization term}, that will force the regression weights to respect certain constraints. A very popular regularizer is the $l_1$-norm of $\betavec$, $\lonenorm{\betavec} = \sum_{p=1}^m |\beta_p|$, which has the effect of shrinking the $\beta_p$ coefficients and setting a large number of them to zero, hence achieving feature selection: the features with zero weights do not enter the model and can hence be rejected. This results in the lasso~\cite{tibshirani1994}, which estimates the regression weights by solving Eq.(\ref{eqn:objective_lasso}). The reason for using the $l_1$-norm, rather than the $l_0$-norm which counts the number of variables that enter the model and hence directly enforces sparsity, is that with the $l_0$-norm the resulting objective function would be non-convex, making its minimization very challenging computationally. 
\begin{equation}
  \label{eqn:objective_lasso} 
  \argmin_{\betavec \in \mathbb{R}^m} \ltwonorm{X \betavec - \yvec}^2 + \lambda \, \lonenorm{\betavec}.
\end{equation}
Here, $\lambda \in \mathbb{R}^+$ is a parameter which controls the balance between the relevance and the regularization terms.

Many other regularizers have been proposed, to satisfy a variety of constraints on the regression weights, and have led to many contributions for the analysis of GWAS data~\cite{wu2009,zhou2010,chen2010,zhao2011,silver2012}. 

\paragraph{Network regularizers} In particular, it is possible to design regularizers that force the features that are assigned non-zero weights to follow a given underlying structure~\cite{huang2011,micchelli2013}. In the context of network-guided biomarker discovery, we will focus on regularizers $\Omega(\betavec)$ that penalize solutions in which the selected features are not connected over a given network.\\
We are now assuming that we have access to a biological network over the biomarkers of interest. 
Such a network can usually be built from a gene interaction network~\cite{azencott2013}.

A first example of such approaches is the Overlapping Group Lasso~\cite{jacob2009}.
Supposing that the $m$ markers are grouped into $r$ groups $\{G_1, G_2, \dots, G_r\}$, which can overlap, we denote by $\vset_{\mathcal{G}}$ the set of $r$-tuples of vector $\vvec = (v_u)_{u=1, \dotsm r}$ such that $v_u$ is non-zero only on features belonging to group $u$. The Overlapping Group Lasso penalty, defined by Eq.(\ref{eqn:penalty_ogl}), induces the choice of weight vectors $\betavec$ that can be decomposed in $r$ weight vectors $\vvec = (v_u)_{u=1, \dotsm r}$ such that some of the $v_u$ are equal to zero. 
This limits the non-zero weights to only some of the groups. 
If each network edge defines a group of two biomarkers, then this method can be applied to network-guided biomarker discovery, where it will encourage the selection of biomarkers belonging to the same group, i.e linked by an edge.
\begin{equation}
  \label{eqn:penalty_ogl} 
  \Omega_{\mbox{ogl}}(\betavec) = \inf_{\vvec \in \vset_{\mathcal{G}}: \sum_{u=1}^r v_u=\betavec} \sum_{u=1}^r \ltwonorm{v_u}.
\end{equation}

Another way to smooth regression weights along the edges of a predefined network, while enforcing sparsity, is a variant of the Generalized Fused Lasso~\cite{tibshirani2005}. The corresponding penatly is given by Eq.(\ref{eqn:penalty_gfl}). The resulting optimization problem is typically solved using proximal methods such as the fast iterative shrinkage-thresholding algorithm (FISTA)~\cite{beck2009}.
While it has not been applied to biomarker discovery to the best of our knowledge,~\cite{xin2014} successfully applied this approach to Alzheimer's disease diagnostic from brain images.
\begin{equation}
  \label{eqn:penalty_gfl} 
  \Omega_{\mbox{gfl}}(\betavec) = \sum_{p \sim q} |\beta_p - \beta_q| + \eta \lonenorm{\betavec}.
\end{equation}

Alternatively, based on work on regularization operators by Smola and Kondor~\cite{smola2003}, Grace~\cite{li2008,li2010} uses a penalty based on the graph Laplacian $L$ of the biological network, which encourages the coefficients $\betavec$ to be smooth on the graph structure. This regularizer is given by Eq.(\ref{eqn:penalty_grace}), and yields a special case of the recently proposed Generalized Elastic Net~\cite{sokolov2016}. It penalizes coefficient vectors $\betavec$ that vary a lot over nodes that are linked in the network. 
The corresponding optimization problem  can be solved through a coordinate descent algorithm~\cite{friedman2007}. Grace was applied to gene-gene networks, but can theoretically be extended to other types of networks of biomarkers; the aGrace variant allows connected features to have effects of opposite directions.
\begin{equation}
  \label{eqn:penalty_grace} 
  \Omega_{\mbox{grace}}(\betavec) = \betavec^{\top} L \betavec = \sum_{p, q} W_{pq} (\beta_p - \beta_q)^2
\end{equation}

These approaches are rather sensitive to the quality of the network they use, and might suffer from bias due to graph misspecification. GOSCAR~\cite{yang2012} was proposed to address this issue, and replaces the term $|\beta_p - \beta_q|$ in Eq.(\ref{eqn:penalty_gfl}) with a non-convex penalty: $\max \left(|\beta_p|, |\beta_q|\right) = \frac{1}{2} \left(|\beta_p + \beta_q| + |\beta_p - \beta_q| \right)$. The authors solve the resulting optimization problem using the alternating direction method of multipliers (ADMM)~\cite{gabay1976,boyd2011}.

Finally, while the previous approaches require to build a network over biomarkers, 
the Graph-Guided Group Lasso~\cite{wang2014} encourages genes connected on the network to be selected in and out of the model together (graph penalty), and biomarkers attached to a given gene to be either selected together or not at all (group penalty). Supposing that the $m$ biomarkers are grouped into $r$ mutually exclusive genes $\{G_1, G_2, \dots, G_r\}$, and calling $\betavec_{G_u}$ the coefficient vector $\betavec$ restricted to its entries in $G_u$, the Graph-Guided Group Lasso penalty is given by Eq.(\ref{eqn:penalty_wang}). As Grace's, this optimization problem can be solved with a coordinate descent algorithm.
\begin{equation}
  \label{eqn:penalty_wang} 
  \Omega_{\mbox{gggl}}(\betavec) = 
  \sum_{u=1}^r \sqrt{|G_u|} \ltwonorm{\betavec_{G_u}} + 
  \eta_1 \lonenorm{\betavec} + 
  \eta_2 \frac{1}{2} \sum_{\substack{p \in G_u, q \in G_v \\ G_u \sim G_v}} W_{uv} (\beta_p - \beta_q)^2.
\end{equation}

\paragraph{Limitations}

In practice, we found that the computational burden was a severe limitation to applying the Overlapping Group Lasso and Grace to the analysis of more than a hundred thousand markers~\cite{azencott2013}. 
On a similar note, the experiments presented in~\cite{yang2012} used at most $8,000$ genes; the graph-guided group lasso~\cite{wang2014} used $1,000$ SNPs only; and the work in~\cite{xin2014} used $3,000$ voxels to describe brain images. It is therefore unclear whether these methods can scale up to several hundreds of thousands of markers.\\
While these computational issues might be addressed by using more powerful solvers or parallel versions of the algorithms, these regularized linear regression approaches also suffer from their inability to guarantee their  {\em stability} as feature selection procedures, meaning their ability to retain the same features upon minor perturbations of the data.
These algorithms are typically highly unstable, often yielding widely different results for different sets of samples relating to the same phenotype~\cite{dernoncourt2014}. 
There is hope that the use of structural regularizers, such as those we defined above, can address this phenomenon by helping the selection of ``true'' features, but ranking features based on t-test scores often still yields the most stable selection in practice~\cite{haury2011,kuncheva2012}.
\\
Finally, it is interesting to note that biomarkers are often represented as categorical variables (such as the presence or absence of a mutation, or the number of minor alleles observed in the case of SNPs). Applying linear (or logistic) regressions in this context, although not entirely meaningless, can be considered an unsatisfying choice.

\subsection{Penalized relevance}
Let us assume data is described  over a set $\vset$ of $m$ features. 
The penalized relevance framework proposes to carry out feature selection by identifying the subset $\sset$ 
of $\vset$ that maximizes the sum of a data-driven {\em relevance function} and a domain-driven {\em regularizer}.

The relevance function $R: 2^\vset \rightarrow \mathbb{R}$ quantifies the importance of a set of features with respect to the task under study. 
It can be derived from a measure of correlation, or a statistical test of association between groups of features and a phenotype.

Our objective is to find the set of features $\sset \subseteq \vset$ that maximizes $R$ under structural constraints, which we model, as previously, by means of a regularizer $\Phi: 2^\vset \rightarrow \mathbb{R}$, which promotes sparsity patterns that are compatible with a priori knowledge about the feature space.
A simple example of regularizer computes the cardinality of the selected set.
More complex regularizers can be defined to enforce a specific structure on $\sset$, and in particular a network structure~\cite{azencott2013}.
We hence want to solve the following problem:
\vspace{-.55em}
\begin{equation}
  \label{eqn:objective_rr} 
  \argmax_{\sset \subseteq \vset} R(S) - \lambda \Phi(S).
\end{equation}
Here again, $\lambda \in \mathbb{R}^+$ is a parameter which controls the balance between the relevance and the regularization terms.

This formulation is close to that of the regularized linear regression presented in Section~\ref{sec:regularized_regression}.
However, Lasso-like approaches focus on the minimization of an empirical risk (or prediction error), while the penalized relevance framework shifts the emphasis to the maximization of feature importance with respect to the question under study.
As with the approaches presented in Section~\ref{sec:post_hoc}, this formulation makes it possible to leverage a large body of work from statistical genetics to define relevance based on appropriate statistical tests.
Moreover, in this framework, optimization is done directly over the power set of $\vset$ (also noted as $2^\vset$), rather than over $\mathbb{R}^m$.
This presents the conceptual advantage of yielding sparsity formulations that can be optimized without resorting to convex relaxation, and offers better computational efficiency in very high dimension.

When relying on linear models, relevance functions are modular, meaning that the relevance of a set of biomarkers is computed as the sum of the relevances of the individual biomarkers in this set.
Moreover, a number of submodular, structure-enforcing regularizers can be derived from sparsity-inducing norms~\cite{bach2010}. 
Among them, the Laplacian-based graph regularizer, which encourages the selected features to be connected on a predefined graph defined by its adjacency matrix $W$, is very similar to $\Omega_{\mbox{grace}}$ in Eq.(\ref{eqn:penalty_grace}). It is  given by 
\begin{equation}
  \label{eq:laplacian_regularizer}
  \Phi_{\mbox{Laplacian}}: \sset \mapsto \sum_{p \in \sset} \sum_{q \notin \sset} W_{pq}.
\end{equation}

The sum of submodular functions is submodular, hence if $R$ is modular and $\Phi$ submodular, 
solving Eq.(\ref{eqn:objective_rr}) becomes a submodular minimization problem and can be solved in polynomial time.
Unfortunately, algorithms to minimize arbitrary submodular functions are slow ($\mathcal{O}(m^5 c + m^6)$ where $c$ is the cost of one function evaluation~\cite{orlin2009}).
However, faster algorithms exist for specific classes of submodular functions. In particular, {\em graph cut functions} can be minimized much more efficiently in practice with maximum flow approaches~\cite{greig1989}, a particularity that has long been exploited in the context of energy minimization in computer vision~\cite{kolmogorov2004}. 

\begin{figure}
  \vspace{-1em}
  \begin{center}
    \includegraphics[width=0.65\textwidth]{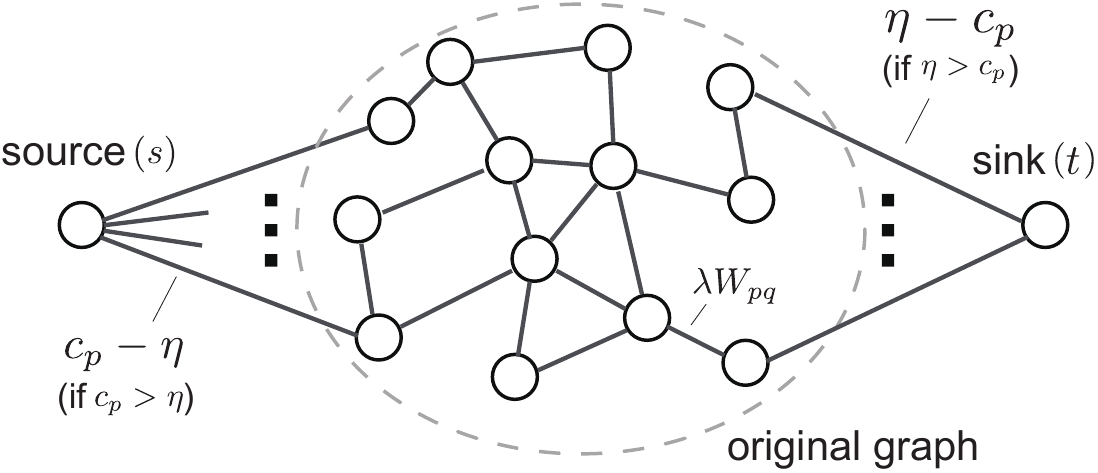}
  \end{center}
  \caption{This figure, taken from~\cite{azencott2013}, shows a graph for which finding the minimum cut is equivalent to maximizing the objective function in Eq.(\ref{eqn:objective_scones}). $c_p$ denotes the relevance of biomarker $p$, and $W_{pq}$ is the weight of the edge connecting biomarker $p$ to biomarker $q$ in the given network.}
  \label{fig:scones_cut}
\end{figure}

This property can be exploited in the specific case of penalized relevance implemented in SConES~\cite{azencott2013}, where $R$ is defined by linear SKAT~\cite{wu2011} and $\Phi$ by the sum of a cardinality constraint $\eta |\sset|$ and the Laplacian-based regularizer
$\Phi_{\mbox{Laplacian}}$ defined above.
 SConES solves the optimization problem given by Eq.(\ref{eqn:objective_scones}):
\vspace{.2em}
\begin{equation}
  \label{eqn:objective_scones} 
  \argmax_{\sset \subseteq \vset} 
  \sum_{p \in \sset}  R(\{p\}) - 
  \eta |\sset|
  - \lambda \sum_{p \in \sset} \sum_{q \notin \sset} W_{pq}.
\end{equation}
In this case, the submodular minimization problem can be cast as a graph-cut problem and solved very efficiently. 
Figure~\ref{fig:scones_cut} shows the transformed $s/t$-graph for which finding a minimum cut is equivalent to solving Eq.(\ref{eqn:objective_scones}).  
This approach is available as a Matlab implementation\footnote{\url{https://github.com/chagaz/scones}} as well as part of the \texttt{sfan} Python package\footnote{\url{https://github.com/chagaz/sfan}}.

\paragraph{Limitations}
An important aspect of both regularized regression and penalized relevance is the parameters (such as $\lambda$ or $\eta$) that control the trade-off between the different terms and regularizers. 
While they afford these methods their flexibility, one needs to come up with appropriate ways to set them. 
This is typically done in an internal cross-validation setting, in which one explores several possible values for each of the parameters, and choose those leading to the best performance according to a given criterion. In biomarker discovery, this criterion can either be the predictivity of the selected biomarkers in a given model, or their stability~\cite{kuncheva2007}.
Finding a good balance between both aspects is difficult, as approaches that either select all or none of the features will have high stability but poor predictivity -- and little interest. 
In addition, exploring multiple parameter values increases the computational cost of these approaches.
\\ 
While SConES is computationally more efficient than the regularized regression approaches, it also suffers from the limitation of relying on an additive model, in which the final phenotype is a function of a linear combination of the individual effects of each biomarker. Biology, however, is highly non-linear, and we expect the effect of a combination of biomarker to be more accurately approached by non-linear models. However, such models lead to optimization problems that are far more computationally expensive to solve.

\subsection{Multi-task extensions}
\paragraph{Multi-task setting} The assumption that there are benefits to be gained from jointly learning on related tasks has long driven the fields of multi-task learning and  multi-task feature selection. 
This also holds for biomarker discovery~\cite{park2011,oreilly2012}
For example, in toxicogenomics, where one studies the response of a population of cell lines to exposure to various chemicals~\cite{eduati2015}, one could try to perform feature selection for each chemical separately, but jointly selecting features for all chemicals reduces the features-to-sample ratio of the data.
eQTL studies, which try to identify the SNPs driving the expression level of various genes, also fall within this setting~\cite{cheng2014}.

\paragraph{Multi-task regularized linear regression}

Many multi-task variants of the lasso have been proposed~\cite{obozinski2006,zhou2010}, and can be extended in spirit to various structural regularizers, such as Grace~\cite{sugiyama2014}.
Assuming $T$ tasks, each containing $n_t$ training samples, and denoting by $\betavec_t$ the $m$-dimensional vector of regression weights for task $t$, the first of these approaches consists in solving the optimization problem defined by Eq.(\ref{eqn:objective_obozinski}). The penalty term used enforces that the regression weights are both sparse and smooth across tasks.
\begin{equation}
  \label{eqn:objective_obozinski} 
  \argmin_{\betavec_1, \dots, \betavec_t \in \mathbb{R}^{m}} \sum_{t=1}^T \frac{1}{n_t} \sum_{i=1}^{n_l} (X_i \betavec_t - y_i)^2 + 
  \lambda \, \sum_{t=1}^T \ltwonorm{\betavec_t}.
\end{equation}

When a network structure is known over the {\em phenotypes}, the graph-fused Lasso can be used to smooth coefficients across tasks~\cite{kim2009}. 
One could imagine combining this approach with a graph regularizer over the features.
Although this has not been done with the graph-fused Lasso, the authors of~\cite{wang_curry2014} successfully used Laplacian-based regularizers both on the biomarkers and on the phenotypes to analyze associations between DNA methylation (about $15,000$ CpG probes) and gene expression. 
In related work, the authors of~\cite{fei2013} use a Laplacian-based penalty to discover the structure of the correlation between the traits.

Most of the multi-task approaches that have been proposed for regularized regression assume that the same features should be selected across all tasks. Indeed, while the multi-task lasso of~\cite{obozinski2006} allows for different regression weights for the selected features, it imposes that the same features have non-zero weights across all tasks.
While this is reasonable for some application domains, this assumption is violated in a number of biomarker discovery settings. For instance, lung diseases such as asthma and chronic obstructive pulmonary disease may be linked to a set of common mutations, but there is no indication that the exact same mutations are causal in both diseases. One way to address this problem is to decompose the regression weights in two components, one that is common to both tasks and one that is task specific, but this increases the computational complexity and is not yet amenable to hundreds of thousands of biomarkers~\cite{swirszcz2012,bellon2016}.

Moreover, to the best of our knowledge, none of the multi-task regularized regression methods that incorporate structured regularizers make it possible to consider different structural constraints for different tasks. However, we may for example want to consider different biological pathways for different diseases.

\paragraph{Multi-task penalized relevance}

Because of the computational efficiency of graph-cut implementations, SConES can be extended to the multi-task setting in such a way as to address these issues. Multi-SConES~\cite{sugiyama2014} proposes a multi-task feature selection coupled with multiple network regularizers to improve feature selection in each task by combining and solving multiple tasks simultaneously.\\
The formulation of Multi-SConES is obtained by the addition of a regularizer across tasks.
Assuming again $T$ tasks, and denoting by $\bigtriangleup$ the symmetric difference between two sets, this formulation is given by Eq.(\ref{eqn:objective_multiscones}).
\begin{equation}
  \label{eqn:objective_multiscones} 
  \argmax_{\sset_1, \dots, \sset_T \subseteq \vset} 
\sum_{t=1}^T \left( \sum_{p \in \sset_t}  R(\{p\}) - \eta |\sset_t|
- \lambda \sum_{p \in \sset_t} \sum_{q \notin \sset_t} W_{pq} \right)
- \mu \sum_{t < u} |\sset_u \bigtriangleup \sset_v|.
\end{equation}

\paragraph{Limitations} The main challenges of current multi-task approaches for biomarker discoveries are linked to their computational complexity, which grows at best linearly with the number of tasks. Allowing different features to be selected across tasks, imposing different network constraints for different tasks, and leveraging prior knowledge on the correlation structure between tasks all increase the computational complexity of the model, which currently limits the applicability of existing methods to a handful of tasks at most.

\section{Open Problems}
\label{sec:open-problems}
The three main challenges in network-guided biomarker discovery today are: departing from linear models; guaranteeing stability; and evaluating the statistical significance of the detected modules.

\paragraph{Problem 1.}\textbf{There is no method that incorporates network information and accounts for non-linear effects between genetic loci. } 
Non-additive epistatic effects are believed to play an important role in a number of human diseases, such as breast cancer~\cite{ritchie2001}, ovarian cancer~\cite{larson2014}, hypertension~\cite{williams2004}, or type-2 diabetes~\cite{cho2004}.\\ 
A large number of methods, reviewed in~\cite{niel2015}, have been proposed to perform exhaustive association tests between {\em pairs} of SNPs and a phenotype. A first step to address the lack of approaches relying on biological networks for the detection of non-linear interaction effects between SNPs and a phenotype would be to combine them with the approaches outlined above; the penalized relevance framework lends itself particularly well to this. However, this is still limited to interactions between two loci, but more might be at play, and models for higher-order interactions are required.\\
Embedded approaches for feature selection are not limited to linear algorithms.
Several promising approaches have been proposed in recent years along those lines, based mostly on random forets~\cite{yoshida2011,stephan2015}, but also on Bayesian neural networks~\cite{beam2014}.\\
Alternatively, Drouin et al.~\cite{drouin2014} propose to use set covering machines~\cite{marchand2002} to learn conjunctions of disjunctions of short genomic sequences to predict bacterial resistance to antibiotics.
Unlike random-forests-based approaches, this approach only consider specific types of biomarkers interactions (combinations of logical ANDs and ORs on their presence/absence), but it also has the potential to uncover epistatic interactions not detectable with the usual quadratic methods.\\
However, to the best of our knowledge, no approach exist that allows for the integration of prior knowledge as networks in these higher-order, non-linear interaction models, and this would be an exciting research avenue to pursue. 

\paragraph{Problem 2.}\textbf{There is no method to guarantee  stable feature selection.} 
The {\em stability} (or {\em robustness}) of feature selection procedures, meaning their ability to retain the same features upon minor perturbations of the data, remains a major predicament in the high-dimensional, low sample-size setting. 
Current algorithms typically yield widely different results for different subsets of the same set of samples~\cite{dernoncourt2014}. 
This high variability implies that they capture idiosyncrasies rather than truly relevant features.
This casts doubts on the reliability of predictive algorithms built on the selected features and impedes interpreting these features to yield novel biological insights.
However, this question has only recently started to come under investigation~\cite{he2010}.\\
Most of the work in that domain has tried to yield lower-dimensional representations by grouping features together in meta-features, based either on the data or on prior knowledge~\cite{ma2009,yu2008}.
Unfortunately, these groupings, if done wrongly, can confuse the feature selection procedure even more.\\
Alternatively, ensemble approaches are based upon the idea of ensemble learning methods to combine the strengths of multiple weak learners to form a stronger predictor. 
Bagging approaches, in which each of the selector is based on a subsample of the data, have been shown to be consistent in settings in which the procedure based on the full data was not~\cite{meinshausen2010,shah2013}.\\
Finally, variable-reduction approaches have led to schemes which reweight samples based on their suitability for the estimation of feature importance~\cite{han_yu2012}. 
However, all these efforts are in their infancy and ranking features based on t-test scores often still yields the most stable selection in practice~\cite{haury2011,kuncheva2012}.

\paragraph{Problem 3.}\textbf{There is no method to assess the statistical significance of the uncovered modules.}
Very few methods can determine the statistical significance of the association between {\em multiple} biomarkers and a phenotype, despite it being key to the interpretation of biomarker discovery outcomes. 
A recent paper~\cite{llinareslopez2015} proposes to do this for intervals of the genome. The extension of this work to network modules, however, is not trivial. \\
Work on confidence intervals on edge differences between brain imaging networks~\cite{belilovsky2016} solve a related problem. However, in the case of network-guided biomarker discovery, one is interested in evaluating the significance of node (and not edge) differences, and it is not obvious whether biological networks can be described with similar models as brain images.

\section{Future Outlook}

We can hardly hope to understand the biology underlying complex diseases without considering the molecular interactions that govern entire cells, tissues or organisms.  
The approaches we discussed offer a principled way to perform biomarker discovery in a systems biology framework, by integrating knowledge accumulated in the form of interaction networks into studies associating genomic features with a disease or response to treatment.
While these methods are still in their infancy, in strong part because of the statistical and computational challenges outlined in Section~\ref{sec:open-problems}, we believe that they can become powerful tools in the realization of precision medicine.

Future research directions for biomarker discovery include the development of (1) machine learning approaches for stable, non-linear, multi-task feature selection;
(2) statistical techniques for the evaluation of the significance of the association detected by complex models;
and (3) the refinement and choice of appropriate network data. 
While most network-guided biomarker discovery studies make use of generic gene-gene interaction networks such as STRING or BioGRID, many other possibilities are starting to open up. They include disease-specific  networks such as ACSN, but we can also imagine using for example eQTL networks based on previous studies~\cite{tur2014}, or three-dimensional chromatin interaction networks~\cite{sandhu2012}. Methods that integrate these multiple types of networks may be needed; that the regularized regression or penalized relevance methods we discussed can all accomodate weighted networks (either directly or through simple modifications) will facilitate these developments.

Finally, serious progress in the field of biomarker discovery requires proper validation, at the very least in other data sets pertaining to the same trait, of the pertinence of the modules identified by these various methods. Because this requires that modelers convince the owners of such data sets to run experiments to this end, this is often hard to implement outside of large consortium collaborations, and a major limitation of most of the work cited in this chapter.  

\bibliographystyle{splncs}
\makeatletter
\renewcommand\@biblabel[1]{#1. }
\makeatother
\bibliography{references}

\end{document}